\documentclass[10pt,twocolumn,letterpaper]{article}

\usepackage{cvpr}              

\usepackage{times}
\usepackage{epsfig}
\usepackage{graphicx}
\usepackage{amsmath}
\usepackage{amssymb}
\usepackage{booktabs}
\usepackage{caption}
\usepackage{color}
\usepackage{multirow}
\usepackage{enumitem}
\usepackage[dvipsnames]{xcolor}
\usepackage[ruled, lined,noend, linesnumbered, commentsnumbered]{algorithm2e}
\usepackage{tikz}
\usepackage{float}
\usepackage{fontawesome5}
\usepackage[accsupp]{axessibility}
\graphicspath{{figs/}}

\newcommand{\mysection}[1]{\vspace{2pt}\noindent\textbf{#1}}

\definecolor{secLock}{HTML}{00b050}
\definecolor{unsecLock}{HTML}{ff0000}
\definecolor{commentsColor}{RGB}{219, 48, 122}

\SetCommentSty{mycommfont}
\let\oldnl\nl
\newcommand{\nonl}{\renewcommand{\nl}{\let\nl\oldnl}}

\usepackage[normalem]{ulem}
\useunder{\uline}{\ul}{}

\usepackage[precision=2, unit=mm]{lengthconvert}
\usetikzlibrary{patterns}

\definecolor{cvprblue}{rgb}{0.21,0.49,0.74}
\usepackage[pagebackref,breaklinks,colorlinks,citecolor=cvprblue]{hyperref}

\def\paperID{6064} 
\def\confName{CVPR}
\def\confYear{2024}

\title{Privacy-preserving Optics for Enhancing Protection in Face De-identification}

\author{Jhon Lopez$^{1,2}$, Carlos Hinojosa$^{2,*}$, Henry Arguello$^{1, \dagger}$, Bernard Ghanem$^{2, \dagger}$\\
$^{1}$Universidad Industrial de Santander; \ \ $^{2}$King Abdullah University of Science and Technology (KAUST) \\
{\tt\small \url{https://carloshinojosa.me/project/privacy-face-deid/}}
}

\begin{document}
\maketitle

\def\thefootnote{*}\footnotetext{Project lead; $\dagger$ Equal PI contribution.}

\begin{abstract}
The modern surge in camera usage alongside widespread computer vision technology applications poses significant privacy and security concerns. Current artificial intelligence (AI) technologies aid in recognizing relevant events and assisting in daily tasks in homes, offices, hospitals, etc. The need to access or process personal information for these purposes raises privacy concerns. While software-level solutions like face de-identification provide a good privacy/utility trade-off, they present vulnerabilities to sniffing attacks. In this paper, we propose a hardware-level face de-identification method to solve this vulnerability. Specifically, our approach first learns an optical encoder along with a regression model to obtain a face heatmap while hiding the face identity from the source image. We also propose an anonymization framework that generates a new face using the privacy-preserving image, face heatmap, and a reference face image from a public dataset as input. We validate our approach with extensive simulations and hardware experiments.
\end{abstract}
\vspace{-15pt}
\section{Introduction}
\label{sec:intro}


The widespread use of cameras, coupled with the ubiquitous application of computer vision technology across various facets of our daily lives, has led to a significant and increasing concern about privacy and security. Specifically, current AI technology enables systems to recognize relevant events and assist us in daily activities in homes, offices, hospitals, etc. However, how can we ensure this technology protects our privacy, especially when it needs to access or process our personal information? This rising concern over data privacy has led international entities like the European Union to establish regulations for ensuring privacy \cite{viorescu20172018}. 



Different studies have explored methods and applications for protecting privacy in computer vision. In general, these studies can be divided into software and hardware-level protection. Most methods operate at the software-level since they directly work on already acquired high-fidelity images. Prior approaches rely on domain knowledge and hand-crafted approaches, such as pixelation, blurring, and face/object replacement, to protect sensitive information \cite{ravi2023review, liu2022privacy, zhao2023visual}. However, traditional anonymization methods, such as face blurring, change the image semantics, resulting in a significant detection performance drop in downstream tasks such as face detection~\cite{Jiang_2023_CVPR} and tracking~\cite{voigtlaender2019mots}. Recent studies have shown that de-identifying people in images, \ie, removing the identification characteristics and generating virtual faces to replace these identities, maintains the utility for downstream tasks~\cite{hukkelaas2019deepprivacy, Maximov_2020_CVPR, wen2023divide, barattin2023attribute}.

\begin{figure}
    \centering
    \includegraphics[width=\columnwidth]{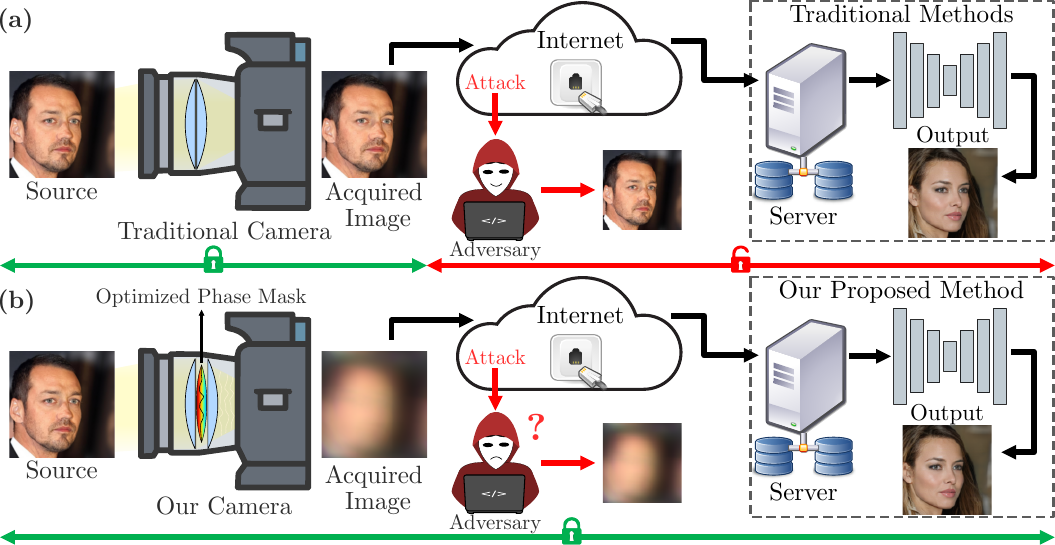}
    \vspace{-10pt}
    \caption{\small \textbf{Traditional \vs our proposed pipeline.} (a) The traditional face de-identification pipeline is vulnerable to adversarial attacks (\textcolor{unsecLock}{\tiny \faIcon{unlock}}). (b) We propose to close the security gap and enhance protection by learning privacy-preserving optics (\textcolor{secLock}{\tiny \faIcon{lock}}).}
    \vspace{-10pt}
    \label{fig:intro}
\end{figure}



Unfortunately, these software-level solutions are vulnerable to adversarial attacks that can get direct access to the original privacy-sensitive images before the vision algorithms process them (see Fig. \ref{fig:intro} (a)). Consequently, hardware-level privacy protection approaches are considered more secure, since they rely on the optical system to add an extra layer of security by removing sensitive data during image acquisition. Prior hardware-level methods include the use of low-resolution cameras \cite{ryoo2018extreme}, defocus lenses \cite{pittaluga2015privacy}, and depth cameras \cite{srivastav2019human} to perform different computer vision tasks like human action recognition and pose estimation. Recent approaches employ the end-to-end and joint optimization of the optics and vision task to promote privacy \cite{hinojosa2021learning,tasneem2022learning}. With such approaches, it is possible to learn a phase mask \cite{sitzmann2018end} to modulate the incident light field to filter out privacy-sensitive information before image capture.

\mysection{Paper Contribution.} Among the privacy-preserving solutions, face de-identification using generative adversarial networks (GAN) provides the best privacy/utility trade-off. However, current face de-identification methods operate at the software-level, introducing a security gap between image acquisition and algorithms (see Fig. \ref{fig:intro} (a)). In this paper, we develop a hardware-level face de-identification approach to close the security gap as shown in Fig. \ref{fig:intro} (b). In our study, we observe that the learned optics filter out high-frequency information from the image to protect privacy and promote face misrecognition; however, low-frequency information is preserved. Therefore, we first perform end-to-end training of our privacy-preserving computational camera (a.k.a optical encoder) and a heatmap regression model to obtain a face heatmap that conceals privacy-sensitive information from the source image. Then, we train a GAN using the acquired privacy-preserving image, the corresponding face heatmap, and a reference face image from a public dataset to generate a new synthetic face. The acquired privacy-preserving image and the face heatmap conceal the true identity of the source but provide general information that allows our model to understand the global geometry (e.g., head position). On the other hand, the given reference image is used to extract the specific style for the new face. We summarize our contributions as follows:
\textbf{(i)} We introduce a full privacy-preserving framework for face de-identification that addresses the security gap between image acquisition and algorithms in traditional approaches. \textbf{(ii)} Our approach consists of two integrated stages: First, we jointly optimize the lens of a computational camera to encode the source images such that the true identity of the person is concealed, but important features are preserved to perform face heatmap regression. Second, we use the acquired privacy-preserving images, the corresponding face heatmaps, and reference images from a public dataset to train a GAN to generate a new face. \textbf{(iii)} We propose a facial expression consistency loss and use an LPIPS-based loss term to improve quality and promote the preservation of detailed expressions (like smiles) in the generated images. \textbf{(iv)} We perform extensive simulations on FFHQ and CelebA-HQ datasets to validate our proposed approach. Furthermore, we build a prototype camera and perform hardware experiments that match simulations.


\section{Related Work}
\label{sec:related_works} 

We categorize previous efforts in privacy-preserving vision into two groups: \textit{software-level} and \textit{hardware-level} protection, where the latter is considered more robust to attacks.

\subsection{Software-level Privacy Protection}

In the literature, most privacy-preserving vision approaches work at the software level. Specifically, such methods only perform software-level processing on high-resolution images acquired by traditional cameras.

\mysection{Non-generative Approaches.} Traditional methods leverage domain knowledge and use hand-crafted techniques such as blurring, mosaicing, masking, pixelation, and face/object replacement to protect sensitive information \cite{orekondy2018connecting, kumawat2022privacy, ravi2023review, liu2022privacy, zhao2023visual, ahmad2023person}. 
These methods can successfully conceal the identity of a person and private sensitive objects on an image when the target to protect is known beforehand. However, they also could introduce artifacts harming the performance of downstream visual tasks \cite{Jiang_2023_CVPR}.


\mysection{GAN-based Face De-identification.} On the other hand, face de-identification using GAN provides a better privacy/utility trade-off than other approaches. These methods focus on generating virtual faces to replace the original identity, preserving privacy and avoiding being recognized by unauthorized users (adversaries/hackers) and computer vision systems. Several methods in the literature \cite{hukkelaas2019deepprivacy, cao2021personalized, wang2021infoscrub, luo2022styleface, li2023riddle, wen2023divide, barattin2023attribute} are built on top of state-of-the-art GAN networks like StyleGAN2 \cite{Karras_2020_CVPR} and StarGAN2 \cite{choi2020stargan} given their impressive ability to capture the distribution of the training samples and then generate good-looking face images. Among prior works, DeepPrivacy \cite{hukkelaas2019deepprivacy} first extracts the sparse facial key points from the image and masks the face region with a constant value; then employs a StyleGAN2 generator to in-paint a randomly generated face, preserving the contextual and pose information. Similarly, CIAGAN \cite{Maximov_2020_CVPR} also masks the source face region and uses a conditional GAN to perform conditional identity swapping by employing an identity discriminator to force the generator to synthesize a new face on the masked region. Most recent approaches aim to improve the ability of the models to generate better quality and natural-looking faces while keeping similar privacy preservation results as CIAGAN and DeepPrivacy \cite{luo2022styleface, barattin2023attribute, wen2023divide, li2023riddle}. For example, Barattin \etal \cite{barattin2023attribute} proposes to directly optimize the latent space to ensure the identity is distant enough from the original image while preserving some facial attributes.
While these software-level approaches preserve privacy in the final application, the acquired images are not protected. This security gap makes these methods vulnerable to adversarial attacks that can get direct access to the original privacy-sensitive image captures before the algorithms process them.

\subsection{Hardware-level Privacy Protection}
These approaches are considered more secure since they rely on the optical system to add an extra layer of security that removes sensitive data during image acquisition. 

\mysection{Fixed Optics.} Prior methods include the use of low-resolution cameras to capture images and videos, avoiding the unwanted leak of identity information from human subjects \cite{ryoo2017privacy, ryoo2018extreme}. Similarly, authors in \cite{pittaluga2015privacy, pittaluga2016pre} propose optical designs based on a defocusing lens to filter or block sensitive information directly from the incident light field before sensor measurement acquisition and perform K-anonymity to protect privacy. In particular, they show how to select a defocus blur that provides a certain level of privacy over a working region within the sensor size limits; however, only using optical defocus for privacy may be susceptible to reverse engineering attacks \cite{hinojosa2021learning,hinojosa2022privhar}. 

\mysection{Learning Optics.} Instead of using fixed optics privacy-preserving cameras to acquire the data and then train algorithms, the privacy/utility trade-off can be improved at the hardware-level by approaches that jointly optimize the optics and vision algorithms to directly filter sensitive data and enhance utility in downstream tasks \cite{sitzmann2018end, hinojosa2021learning}. Most recent approaches learn a phase mask \cite{hinojosa2021learning, wang2023deep} to modulate the incident light field and filter out privacy-sensitive information before image acquisition. This idea has been applied in different tasks, such as human pose estimation \cite{hinojosa2021learning}, human action recognition \cite{hinojosa2022privhar}, image captioning \cite{arguello2022optics}, and passive depth estimation \cite{tasneem2022learning}. 
Our proposed framework, developed in the next section, uses privacy-preserving cameras to improve privacy and close the security gap while preserving the utility given by face de-identification methods. To the best of our knowledge, this is the first work that learns optics for the face de-identification task. Note that our framework can be used with fixed optics cameras, e.g., low-resolution cameras; however, learning the optics provides better control over privacy and utility, as we show in Section \ref{sec:experiments}.
\section{Proposed Methodology}
\label{sec:method}


\begin{figure*}
    \centering
    \includegraphics[width=\textwidth]{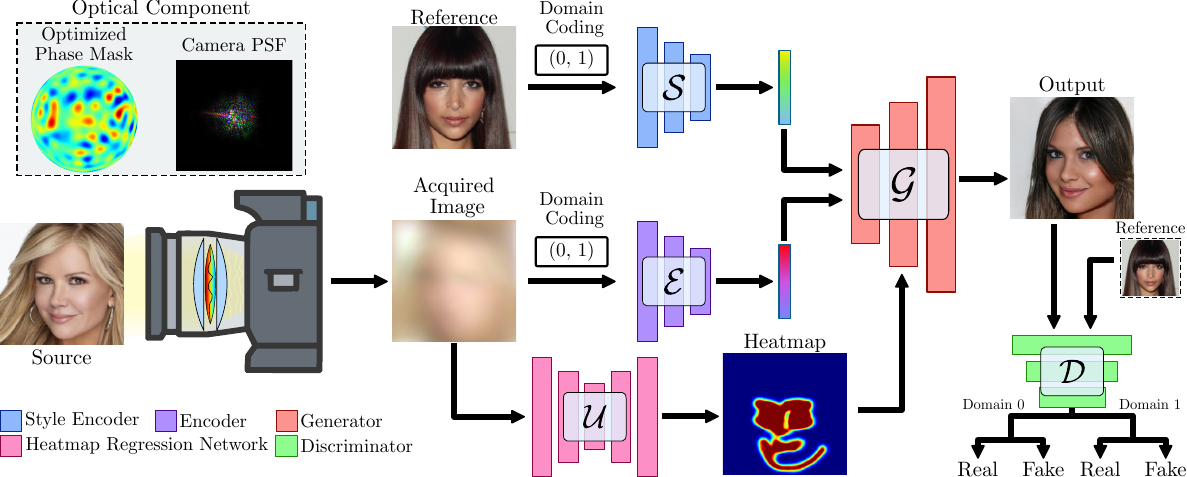}
    \vspace{-15pt}
    \caption{\small \textbf{Our proposed face de-identification framework}. We first jointly optimize the camera optics and a heatmap regression network to obtain a face heatmap and conceal source image identity. Then, we train a GAN network to generate a new face identity using the acquired image, the obtained heatmap, and a reference image. We leverage global geometry information (e.g. face pose and eyes, nose, mouth position) from the acquired image and the heatmap while the style is extracted from the reference image.}
    \vspace{-15pt}
    \label{fig:method}
\end{figure*}



We are interested in using privacy-preserving images to add an extra layer of security to the face de-identification task. Previous approaches \cite{hinojosa2021learning, hinojosa2022privhar, arguello2022optics} learn optics for a specific task, and the utility of the acquired privacy-preserving images is limited only to this task. To overcome this limitation, we propose to combine optics learning with a generative model to maximize utility in downstream tasks. Specifically, the output of our framework is a realistic face image that maintains the anonymity of the original person. 

In general, our framework (Fig. \ref{fig:method}) consists of two parts: optics learning and new face generation. In the first part, our strategy is to jointly optimize the camera optics and a heatmap regression network to obtain a face heatmap while ensuring privacy protection. Our key observation is that we can learn the camera lens to degrade the image such that the high-frequency information is filtered out, concealing the subject identity while preserving important features to obtain a face heatmap. We can then leverage the face heatmap to infer the global geometry information of the original (source) image like head pose and eyes, nose, and mouth position. In the second part, our main goal is to learn the parameters of a conditional GAN to generate a new face identity from the input privacy-preserving image, the face heatmap, and a reference image.

\subsection{Optics Learning}

The main goal of the optical component in our proposed approach (Fig. \ref{fig:method}) is to design a phase mask to visually distort images by removing the high-frequency spatial information and hence conceal privacy-sensitive attributes while preserving useful information to extract a face heatmap. 

\noindent\textbf{Image Formation Model and Phase Mask Learning.} We adopt a similar strategy as previous works in \cite{sitzmann2018end, hinojosa2021learning,hinojosa2022privhar} to couple the modeling and design of two essential operators in the imaging system: wave propagation and phase modulation. We model the image acquisition process by defining the point spread function (PSF) in terms of the lens surface profile. This allows emulating the propagation of the wavefront and training the parameters of the refractive lens. Considering the Fresnel approximation and paraxial regime \cite{goodman2005introduction} for incoherent illumination, we describe the PSF as:
{\small
	 \setlength{\belowdisplayskip}{3pt}
	 \setlength{\abovedisplayskip}{3pt}
    \setlength{\arraycolsep}{2pt}
	\begin{eqnarray}
		H(u',v') = |\mathcal{F}^{-1}\{\mathcal{F}\{P(u,v)\cdot W(u,v)\}\cdot T(f_u,f_v)\}|^2,
		\label{eq:psf_continuous}
	\end{eqnarray}
} 

\noindent where $P(u,v)=P_t(u,v)\cdot P_\phi(u,v)$, $W(u,v)$ is the incoming wavefront, $T(\cdot)$ represents the transfer function with $(f_u,f_v)$ as the spatial frequencies, $P_\phi(u,v)=\exp(-ik\phi(u,v))$ with $\phi(u,v)$ as the phase mask and $k=2\pi/\lambda$ as the wavenumber, $P_t(u,v)=\exp \left( -i\frac{k}{2d}(u^2+v^2) \right)$ denotes the light wave propagation phase with $d$ as the object-lens distance, $\mathcal{F}\{\cdot\}$ denotes the 2D Fourier transform, and $(u',v')$ is the spatial coordinate on the camera plane. The values of $\phi(\cdot)$ are modeled via  Zernike polynomials with $\phi(u,v)=\mathcal{R}(\sqrt{u^2+v^2})\cdot\cos{(\arctan{(v/u)})}$, where $\mathcal{R}\left(\cdot\right)$ represents the radial polynomial function \cite{lakshminarayanan2011zernike}. To train the phase mask values using our model, we discretize the phase mask $\phi(\cdot)$ as:
{\small 
	\setlength{\belowdisplayskip}{2pt}
	\setlength{\abovedisplayskip}{2pt}
	\begin{equation}
		\boldsymbol{\phi}=\sum_{j=1}^{p} \beta_j \mathbf{Q}_j,
		\label{eq:zernike_model}
	\end{equation}
}where $\mathbf{Q}_j$ denotes the $j$-th Zernike polynomial in Noll notation, and $\beta_j$ is the corresponding coefficient \cite{born2013principles}. Each Zernike polynomial describes a wavefront aberration \cite{lakshminarayanan2011zernike}; hence the phase mask $\boldsymbol{\phi}$ is formed by the linear combination of $p$ aberrations. In this regard, the optical element parameterized by $\boldsymbol{\phi}$ can be seen as an optical encoder, where the coefficients $\boldsymbol{\beta}=\left\{\beta_j \right\}_{j=1}^p$ determine the data transformation. Therefore, our end-to-end training finds the coefficients $\boldsymbol{\beta}$ that provide the maximum visual distortion of the scene but allow extracting relevant features to obtain a face heatmap via a regression network.

The sensing process of our camera can be modeled as a shift-invariant convolution operation between the source image (scene) and the PSF as:
{\small
	\setlength{\belowdisplayskip}{3pt}
	\setlength{\abovedisplayskip}{3pt}
	\begin{equation}
		\mathbf{y} =\mathcal{C}\left(\mathbf{H}*\mathbf{x}\right) + \boldsymbol{\eta},
    \label{eq:sensing}
	\end{equation}
}where $\mathbf{x}$ represents the discrete color source image of the person we want to protect; $\mathbf{H}$ denotes the discretized version of the PSF \cite{goodman2005introduction} in Eq. (\ref{eq:psf_continuous}); $\boldsymbol{\eta}$ represents Gaussian noise in the sensor, and $\mathcal{C}(\cdot)$ is the camera response function, which we assume linear. Please refer to our supplementary for more details on our light propagation model.

\mysection{Heatmap Regression Network.} We adopt the same implementation as in \cite{choi2020stargan}, which is an adaptation from the Face Alignment Network (FAN) \cite{bulat2017far} and the adaptive wing-based heatmap \cite{Wang_2019_ICCV}. This network consists of stacked Hourglass (HG) \cite{newell2016stacked} layers and coordinate convolutions (CoordConv) \cite{liu2018intriguing}, which encode coordinate information as additional channels before the convolution operation. Given an image $\mathbf{y}$ captured by our camera, we use the regression network $\mathcal{U}$ to obtain a heatmap $\mathbf{m}=\mathcal{U}(\mathbf{y})$ as shown in Fig. \ref{fig:method}. This heatmap fits the eyes, nose, mouth, and chin region and retrieves geometry information about the face. 



\subsection{Generative Network} \label{Sec_3_2}

To perform the face image generation, we use the generator, style encoder, and discriminator network architectures from StarGAN2 \cite{choi2020stargan}. Considering $\omega \in \Omega$ as an arbitrary domain and given a privacy-preserving image $\mathbf{y}$ captured by our camera as in Eq. \ref{eq:sensing}, we first use an inversion encoder $\mathcal{E}$ to map $\mathbf{y}$ to the latent space $\bar{\mathbf{y}}=\mathcal{E}(\mathbf{y}$). Then, we want the generator $\mathcal{G}$ to translate $\mathbf{y}$ into a new image $\mathcal{G}(\bar{\mathbf{y}}, \mathbf{s}, \mathbf{m})$ reflecting a domain-specific style code $\mathbf{s}$ provided by the style encoder $\mathcal{S}$ and the pose and face geometry information provided by the heatmap $\mathbf{m}$. We extract the style code $\mathbf{s}=\mathcal{S}_{\omega}(\mathbf{r})$ from a publicly available reference image $\mathbf{r}$ corresponding to a specific domain $\omega$. Note that using different reference images produces different style codes $\mathbf{s}$. Specifically,  $\mathbf{s}$ represents the style of a particular reference image $\mathbf{r}$ and domain $\omega$; hence it is not necessary to directly provide $\omega$ to the generator $\mathcal{G}$ and allows $\mathcal{G}$ to synthesize images of all domains reflecting the style of $\mathbf{r}$. Finally, the discriminator $\mathcal{D}$ is a multi-task discriminator \cite{Liu_2019_ICCV, pmlr-v80-mescheder18a}, which consists of multiple output branches. Each branch $\mathcal{D}_{\omega}$ learns a binary classification determining whether an image $\mathbf{r}$ is a real image of its domain $\omega$ or a fake image $\mathcal{G}(\bar{\mathbf{y}}, \mathbf{s}, \mathbf{m})$ produced by $\mathcal{G}$. Note that, unlike StarGAN2, we use the reference image $\mathbf{r}$ in the discriminator instead of the source image $\mathbf{x}$ to avoid privacy information leakage.

\subsection{Training Objectives}

We divide our training into two stages. In the first stage, we learn the coefficients of Zernike polynomials $\boldsymbol{\beta}$ and the parameters of the heatmap regression network to produce images that visually conceal the identity of the person while allowing to extract a face heatmap. For the second stage, we freeze the camera and regression model and only train the generative network to translate the acquired privacy-preserving image to a new face using the extracted heatmap and a reference face image.

\vspace{-10pt}
\subsubsection{Stage I: Optical Design}
\vspace{-5pt}
\label{sec:stageI}

This stage aims to extract the face heatmap information from $\mathbf{y}$ by learning the optics. As the starting point of our training, we assume an aberration-free lens and use a pre-trained regression model $\mathcal{U}$.

\mysection{Optics Loss Function.}
To encourage image degradation, we minimize the difference of the acquired image $\mathbf{y}$ with the original image $\mathbf{x}$. Also, for easy and faster convergence, we promote symmetric and low frequencies using a predefined defocus PSF $\mathbf{H}_f$ as a regularizer. Specifically, assuming we acquire $\mathbf{y}$ using Eq. \ref{eq:sensing} and the images are normalized between $\left[0,1\right]$, we define the loss function for our camera lens optimization as:
{\small
	\setlength{\belowdisplayskip}{2pt}
	\setlength{\abovedisplayskip}{2pt}
	\begin{equation}
	    \mathcal{L}_{optics}= \mathbb{E}_{\mathbf{x}}\left[1 - \|\mathbf{x} - \mathbf{y}\|_2^2 
	    + \alpha_1\|\mathbf{H} - \mathbf{H}_f \|_2 \right],
     \label{eq:loss_lens}
	\end{equation}
} where $\| \cdot \|_2^2$ is the standard mean squared error (MSE), and $\alpha_1$ is a hyperparameter. Note that $\mathbf{H}_f$ is a predefined PSF whose parameters are frozen during training. See our supplementary document for details on our regularizer $\mathbf{H}_f$.



\mysection{Heatmap Regression Loss Function}. After initializing $\mathcal{U}$ with the pre-trained weights \cite{Wang_2019_ICCV}, we aim to obtain a heatmap from $\mathbf{y}$ that closely resembles the heatmap obtained from $\mathbf{x}$; therefore, we use the following loss:
{\small
	\setlength{\belowdisplayskip}{3pt}
	\setlength{\abovedisplayskip}{3pt}
	\begin{equation}
	    \mathcal{L}_{hmap}=\mathbb{E}_{\mathbf{x}} \left[ \|\mathcal{U}(\mathbf{y})-\mathcal{U}^{*}(\mathbf{x})\|_1 \right],
	\end{equation}
} \noindent where we use the heatmap $\mathcal{U}^{*}(\mathbf{x})$ as ground truth and $\mathcal{U}^{*}$ is a freeze pre-trained regression model.

\mysection{Full Objective.} We finetune the camera model and the regression network $\mathcal{U}$ end-to-end to gradually distort the optics and learn to extract the heatmap from the $\mathbf{y}$ using the following full objective function:
{\small
	\setlength{\belowdisplayskip}{3pt}
	\setlength{\abovedisplayskip}{3pt}
	\begin{equation}
	    \min_{\mathcal{C}, \mathcal{U}} \ \mathcal{L}_{optics} + \alpha_2 \mathcal{L}_{hmap},
	\end{equation}
} \noindent where $\alpha_2$ is a hyperparameter to control the similarity between $\mathbf{H} \text{ and } \mathbf{H}_f $.

\vspace{-10pt}
\subsubsection{Stage II: Generative Network}
\vspace{-5pt}
Once we train the lens and the regression network parameters, we now train our generative network using the following objectives. Note that the first four losses are adopted from StarGAN2 \cite{choi2020stargan}, while the last two losses are specifically proposed to improve the generation when using our privacy-preserving image $\mathbf{y}$. 

\mysection{Adversarial Objective}. During training, we randomly select a reference image $\mathbf{r}$ and a target domain $\tilde{\omega} \in \Omega$, from which we compute a target style code $\tilde{\mathbf{s}} = \mathcal{S}_{\tilde{\omega}}(\mathbf{r})$. The generator $\mathcal{G}$ receives the latent code $\bar{\mathbf{y}}=\mathcal{E}(\mathbf{y})$ and the mask $\mathbf{m}=\mathcal{U}(\mathbf{y})$ extracted from the image $\mathbf{y}$ and the style code $\tilde{\mathbf{s}}$ as inputs and is trained to produce an output image $\mathcal{G}(\bar{\mathbf{y}},\mathbf{m}, \tilde{\mathbf{s}})$, guided by an adversarial loss function:
{\small
	\setlength{\belowdisplayskip}{3pt}
	\setlength{\abovedisplayskip}{3pt}
\begin{align}
    \mathcal{L}_{adv} = & \ \mathbb{E}_{\mathbf{y}, \omega} \left[ \log \mathcal{D}_{\omega}(\mathbf{r}) \right] \nonumber \\
    & +\mathbb{E}_{\mathbf{y},\tilde{\omega}, \mathbf{r}} \left[ \log (1-\mathcal{D}_{\tilde{\omega}}(\mathcal{G}(\bar{\mathbf{y}},\mathbf{m}, \tilde{\mathbf{s}}))) \right],
\end{align}
} where $\mathcal{D}_{\omega}(\cdot)$ denotes the output of the discriminator $\mathcal{D}$ corresponding to the domain $\omega$.


\mysection{Style Reconstruction}. To ensure that the generator $\mathcal{G}$ effectively incorporates the style code $\tilde{\mathbf{s}}$ in the image generation process, we employ a style reconstruction loss:
{\small
	\setlength{\belowdisplayskip}{3pt}
	\setlength{\abovedisplayskip}{3pt}
\begin{equation}
    \mathcal{L}_{sty} = \mathbb{E}_{\mathbf{y}, \tilde{\omega}, \mathbf{r}} \left[ \| \tilde{\mathbf{s}} - \mathcal{S}_{\tilde{\omega}}(\mathcal{G}(\bar{\mathbf{y}},\mathbf{m}, \tilde{\mathbf{s}})) \|_{1} \right].
\end{equation}
}


\mysection{Style Diversification.} To further enable the generator to produce diverse images, we explicitly regularize $\mathcal{G}$ with the diversity sensitive loss \cite{mao2019mode}:
{\small
	\setlength{\belowdisplayskip}{3pt}
	\setlength{\abovedisplayskip}{3pt}
\begin{equation}
    \mathcal{L}_{ds} = \mathbb{E}_{\mathbf{y},\tilde{\omega},\mathbf{r}_1, \mathbf{r}_2} \left[ \| \mathcal{G}(\bar{\mathbf{y}},\mathbf{m}, \tilde{\mathbf{s}}_1) - \mathcal{G}(\bar{\mathbf{y}},\mathbf{m}, \tilde{\mathbf{s}}_2) \|_{1} \right],
\end{equation}
} where the target style codes $\tilde{\mathbf{s}}_1$ and $\tilde{\mathbf{s}}_2$ are produced by $\mathcal{S}$ using two reference images $\mathbf{r}_1$ and $\mathbf{r}_2$. 


\mysection{Preserving Source Characteristics.} 
To ensure that the domain-invariant attributes (such as pose) of the input source image $\mathbf{x}$ are accurately maintained in the generated image $G(\bar{\mathbf{y}},\mathbf{m}, \tilde{\mathbf{s}})$, we utilize the cycle consistency loss that leverages the information from the mask extracted from the privacy-preserving image $\mathbf{y}$.
{\small
	\setlength{\belowdisplayskip}{3pt}
	\setlength{\abovedisplayskip}{3pt}
\begin{equation}
    \mathcal{L}_{cyc} = \mathbb{E}_{\mathbf{y}, \omega, \tilde{\omega}, \mathbf{r}} \left[ \|\mathbf{y} -\mathcal{G}(\hat{\mathbf{y}}, \hat{\mathbf{m}}, \hat{\mathbf{s}}) \|_1 \right],
\end{equation}
} where $\hat{\mathbf{y}}=\mathcal{E}(\mathcal{G}(\bar{\mathbf{y}},\mathbf{m}, \tilde{\mathbf{s}}))$, $\hat{\mathbf{m}}=\mathcal{U}^{*}(\mathcal{G}(\bar{\mathbf{y}},\mathbf{m}, \tilde{\mathbf{s}}))$, $\hat{\mathbf{s}} = \mathcal{S}_{\omega}(\mathbf{y})$ is the estimated style code of the input privacy-preserving image $\mathbf{y}$, and $\omega$ corresponds to the original domain of $\mathbf{y}$, which is the same as the source image $\mathbf{x}$. 


\mysection{LPIPS.}
We employ an LPIPS loss term \cite{zhang2018unreasonable} to improve the generated image quality and more effectively reflect the reference image face attributes.
{\small
	\setlength{\belowdisplayskip}{3pt}
	\setlength{\abovedisplayskip}{3pt}
\begin{equation}
    \mathcal{L}_{LPIPS} = \mathbb{E}_{\mathbf{y, \tilde{\omega}, \mathbf{r}}}\left[\|\mathcal{Q}(\mathbf{r})-\mathcal{Q}(\mathcal{G}(\bar{\mathbf{y}},\mathbf{m}, \tilde{\mathbf{s}})) \|_2 \right],
\end{equation}
} where $\mathcal{Q}$ is the pre-trained perceptual feature extractor.

\mysection{Facial Expression Consistency.} Since the high-frequency information is missing in the privacy-preserving image $\mathbf{y}$, it is challenging to preserve small expressions like smiles, eyes blinking, or small mouth movements in the generated image. Therefore, we propose a facial expression consistency loss as follows:
{\small
	\setlength{\belowdisplayskip}{3pt}
	\setlength{\abovedisplayskip}{3pt}
\begin{equation}
\mathcal{L}_{expr} = \mathbb{E}_{\mathbf{x}}\left[\| \mathbf{m}^{*}\odot\mathbf{x} - \mathbf{m}\odot\mathbf{y}\|_1 \right],
\end{equation}
} where $\odot$ denotes the element-wise product, and $\mathbf{m}^{*}=\mathcal{U}^{*}(\mathbf{x})$ is the heatmap obtained from the source image $\mathbf{x}$.

\mysection{Full Objective.} Our full objective function can be summarized as follows:
{\small
	\setlength{\belowdisplayskip}{3pt}
	\setlength{\abovedisplayskip}{3pt}
\begin{align}
    \min_{\mathcal{G}, \mathcal{S}, \mathcal{E}} \max_{\mathcal{D}} \Big( & \mathcal{L}_{adv} + \lambda_{sty}\mathcal{L}_{sty} - \lambda_{ds}\mathcal{L}_{ds} + \lambda_{cyc}\mathcal{L}_{cyc} \nonumber \\
    & + \lambda_{LPIPS}\mathcal{L}_{LPIPS} + \lambda_{expr}\mathcal{L}_{expr} \Big),
\end{align}
} where $\lambda_{sty}, \lambda_{ds}$, $\lambda_{cyc}$, $\lambda_{LPIPS}$, and $\lambda_{expr}$ are hyperparameters for each term. See our supplementary document for training details of Stage I and Stage II, and results when using latent vectors instead of reference images, as used in StarGAN2, to generate the style codes.

\vspace{-5pt}
\section{Experimental Results}
\label{sec:experiments}

We extensively evaluate the performance of our proposed face de-identification approach using standard metrics and evaluation protocols and compare it with other methods. 

\mysection{Dataset.} We train our proposed model on two widely used datasets: CelebA-HQ \cite{karras2017progressive} and FFHQ \cite{karras2019style}.  We separate the images of the CelebA-HQ dataset into two domains (male and female) as in \cite{choi2020stargan}. We show our main results and ablation studies with the CelebA-HQ dataset using the same train/eval set split provided by \cite{choi2020stargan}. To quantitatively compare with other methods, we follow the common practice of using the FFQH dataset for training and the CelebA-HQ dataset for evaluation \cite{li2023riddle, wen2023divide}. For a fair comparison, all images are resized to $256 \times 256$ resolution for training, which is the same resolution used in the baselines.

%

\mysection{Metrics.} Following previous works \cite{cao2021personalized, wen2023divide}, we quantitatively evaluate the identity protection given by our face de-identification approach by measuring the $\ell_2$ distance (DIS) of embedding vectors from the de-identified and original faces extracted by a pre-trained face recognition model. Specifically, we employ the Face Recognition library (FR)~\cite{geitgey_face_recognition} and FaceNet \cite{schroff2015facenet}, which is pre-trained on two public datasets: CASIA-Webface \cite{yi2014learning} and VGGFace2 \cite{cao2018vggface2}. We further evaluate the utility ($\uparrow$) of our generated images and the preservation of facial attributes from the source image ($\downarrow$), such as face orientation, lips position, and other attributes, using the Dlib library \cite{kazemi2014one} and MtCNN network \cite{zhang2016joint}. Specifically, we estimate the facial landmarks between the original and de-identified faces and measure their similarity using the $\ell_1$ norm. Additionally, we estimate the facial bounding box to evaluate whether the face is located in the correct position and size. Finally, we use Fréchet inception distance (FID $\downarrow$) to measure the distance between the generated and real distribution. Since our proposed approach uses reference-guided synthesis, each source image is transformed using 10 reference images randomly sampled from the test set and we report the average value.

\subsection{Ablation Studies}

\begin{table}[t!]
\centering
\resizebox{\columnwidth}{!}{%
\setlength{\tabcolsep}{4pt}
\begin{tabular}{ccccccccccc}
\hline
\multicolumn{3}{c}{Components} & \multicolumn{3}{c}{DIS$\uparrow$}                                                                   & \multicolumn{2}{c}{Landmarks $\downarrow$}                        & \multicolumn{2}{c}{Bounding Box $\downarrow$}                     & \multirow{2}{*}{FID$\downarrow$} \\
H             & L    & E   & FR                              & CASIA                           & VGGFace2                        & MtCNN                           & Dlib                            & MtCNN                           & Dlib                            &                                  \\ \hline
\textcolor{red}{\texttimes} & \textcolor{red}{\texttimes} & \textcolor{red}{\texttimes}   & \textbf{0.851} & 1.132                           & 1.313                           & 5.434                           & 6.473                           & 13.667                          & 8.008                           & 34.388                           \\
\textcolor{green}{\checkmark}     & \textcolor{red}{\texttimes}      &  \textcolor{red}{\texttimes}    & {\ul 0.847}    & {\ul 1.162}    & {\ul 1.350}    & 5.309                           & 6.260                           & 12.988                          & 7.654                           & 36.056                           \\
\textcolor{green}{\checkmark}    & \textcolor{green}{\checkmark}     & \textcolor{red}{\texttimes}    & {\ul 0.847}    & \textbf{1.174} & \textbf{1.359} & 4.692                           & 5.683                           & 11.818                          & 7.037                           & 29.338                           \\
\textcolor{green}{\checkmark}  & \textcolor{red}{\texttimes}      & \textcolor{green}{\checkmark}     & 0.664                           & 1.037                           & 1.199                           & \textbf{1.884} & \textbf{2.081} & \textbf{4.694} & \textbf{4.365} & {\ul 24.998}    \\
\textcolor{green}{\checkmark}  &  \textcolor{green}{\checkmark}  & \textcolor{green}{\checkmark}  & 0.697                           & 1.073                           & 1.255                           & {\ul 2.022}    & {\ul 2.259}    & {\ul 4.727}    & {\ul 4.985}    & \textbf{24.972} \\ \hline
\end{tabular}%
}
 \vspace{-5pt}
\caption{\textbf{Quantitative results of the ablation studies}. The best result for each pre-trained face recognition model is in bold, and the second-best result is underlined.}
\label{tab:ablations}
 \vspace{-15pt}
\end{table}

We conducted a series of experiments to investigate the contribution of using the face heatmap (H), $\mathcal{L}_{LPIPS}$ (L), and $\mathcal{L}_{expr}$ (E) loss functions on the performance of our method. Table \ref{tab:ablations} shows the results of our ablation studies. In the table, a checkmark (\textcolor{green}{\checkmark}) denotes the inclusion, and a red cross (\textcolor{red}{\texttimes}) denotes the exclusion of specific components. From the table, we observe that there is an improvement in all metrics when face heatmaps (H \textcolor{green}{\checkmark}) are used. This supports our hypothesis that integrating a face heatmap in our framework effectively maintains geometric information of the face (such as head pose) from the source image. Since we obtain the face heatmap directly from the privacy-preserving image, the source image privacy is still preserved. Similarly, we observe that the inclusion of the $\mathcal{L}_{LPIPS}$ (L \textcolor{green}{\checkmark}) and $\mathcal{L}_{expr}$ (E \textcolor{green}{\checkmark}) significantly improves all the metrics, especially the FID and facial Landmarks, demonstrating their contribution to our framework to preserve detailed expressions (like smiles) in the generated images. Note that we input the reference image instead of the source image to our discriminator. We conducted experiments using the source image instead of the reference in the discriminator. The results show an improvement in the FID; however, the face anonymization metrics (DIS) decrease significantly, which suggests that the face information from the source image is leaked to the generator. Hence, we only use the reference image in our discriminator to maximize privacy preservation. See our supplementary document for results when using the source image in the discriminator.

\subsection{Comparison with other methods}

\begin{table}[t]
\centering
\resizebox{\columnwidth}{!}{%
\setlength{\tabcolsep}{3pt}
\begin{tabular}{cccccccc}
\hline
Method              & DeepPrivacy & CIAGAN      & FIT   & IDeudemon      & RiDDLE & Ours-LR       & Ours  \\ \hline
FR $\uparrow$       & 0.783       & 0.671       & 0.812 & \textbf{0.819} & 0.776  & {\ul 0.795}    & 0.734       \\
CASIA $\uparrow$    & 1.091       & {\ul 0.919} & 1.207 & \textbf{1.228} & 1.033  & 1.123          & 1.071       \\
VGGFace2 $\uparrow$ & 1.187       & 1.085       & 1.224 & 1.233          & 1.129  & \textbf{1.294} & {\ul 1.251} \\ \hline
\end{tabular}%
}
\vspace{-5pt}
\caption{\small \textbf{Quantitative Results - DIS}. We compare our method against SOTA face de-identification methods on the CelebA-HQ dataset using the DIS metrics (FR, CASIA, VGGFace2).}
\label{tab:face_net}
\vspace{-8pt}
\end{table}

\begin{table}[t]
\centering
\resizebox{\columnwidth}{!}{%
\setlength{\tabcolsep}{4pt}
\begin{tabular}{cccccccc}
\hline
\multicolumn{2}{c}{Method}                                                                   & CIAGAN & FIT            & DeepPrivacy    & RiDDLE          & Ours-LR       & Ours     \\ \hline
\multicolumn{2}{c|}{FID $\downarrow$}                                                        & 32.611 & 30.331         & {\ul 23.713}   & \textbf{15.389} & 34.161         & 29.218         \\ \hline
\multicolumn{1}{c|}{\multirow{2}{*}{FD $\uparrow$}} & \multicolumn{1}{c|}{MtCNN} & 0.992  & \textbf{1.000} & \textbf{1.000} & \textbf{1.000}  & \textbf{1.000} & \textbf{1.000} \\
\multicolumn{1}{c|}{}                                           & \multicolumn{1}{c|}{Dlib}  & 0.937  & 0.984          & 0.980          & {\ul 0.991}     & \textbf{0.994} & {\ul 0.991}    \\ \hline
\multicolumn{1}{c|}{\multirow{2}{*}{BB $\downarrow$}} & \multicolumn{1}{c|}{MtCNN} & 20.387 & 7.879          & {\ul 4.654}    & \textbf{3. 824} & 5.358          & 5.314          \\
\multicolumn{1}{c|}{}                                           & \multicolumn{1}{c|}{Dlib}  & 15.476 & 4.218          & {\ul 2.685}    & \textbf{1.700}  & 4.694          & 4.821          \\ \hline
\multicolumn{1}{c|}{\multirow{2}{*}{LM $\downarrow$}}     & \multicolumn{1}{c|}{MtCNN} & 8.042  & 3.572          & 3.280          & \textbf{1.674}  & 2.392          & {\ul 2.298}    \\
\multicolumn{1}{c|}{}                                           & \multicolumn{1}{c|}{Dlib}  & 8.930  & 4.047          & 2.896          & \textbf{1.512}  & 2.749          & {\ul 2.716}    \\ \hline
\end{tabular}%
}
\vspace{-5pt}
\caption{\textbf{Quantitative Results - Detection.} We compare our method against SOTA face de-identification methods on the CelebA-HQ dataset using the Face Detection (FD), Bounding Box (BB), and Landmarks (LM) metrics.}
\vspace{-10pt}
\label{tab:rstl_celeba}
\end{table}

\mysection{Quantitative Results.} We quantitatively compare our framework with several state-of-the-art (SOTA) face de-identification approaches, including: CIAGAN~\cite{maximov2020ciagan}, Face Identity Transformer (FIT) \cite{gu2020password}, DeepPrivacy~\cite{hukkelaas2019deepprivacy}, IDeudemon~\cite{wen2023divide}, and RiDDLE~\cite{li2023riddle}. We also compare the performance of our proposed framework when using a low-resolution (LR) camera instead of learning the lens. Specifically, we simulate an LR camera (fixed optics parameters) and perform the two-stage training described in section \ref{sec:method} without using the Eq. \ref{eq:loss_lens}. Table \ref{tab:face_net} reports the results for all the methods evaluated on the CelebA-HQ dataset using the $\ell_2$ distance (DIS) metrics. As observed, our approach achieved the best performance in the VGGFace2 metric and a similar performance in the other metrics. Also, Ours approach using a low-resolution camera (Ours-LR) performs better than our learned lens (Ours); however, LR is susceptible to reverse engineering attacks as shown in section \ref{sec:deconv}.

Additionally, we also compare our method against SOTA using face detection (FD), bounding box detection (BB), and landmark (LM) metrics in Table \ref{tab:rstl_celeba}. The results show that our approach does not outperform some SOTA methods, such as RiDDLE. The main reason is the significant loss of high-frequency information on the acquired privacy-preserving images, which makes it challenging to generate images that preserve detailed expressions and face geometry from the source image. Although including our proposed loss functions and using heatmaps helps mitigate this problem, a trade-off exists between achieved image degradation and the generative capabilities of our framework. However, even with such loss of information, our method outperforms other methods, such as CIAGAN and FIT.

\begin{figure}[!t]
    \centering
    \includegraphics[width=0.95\columnwidth]{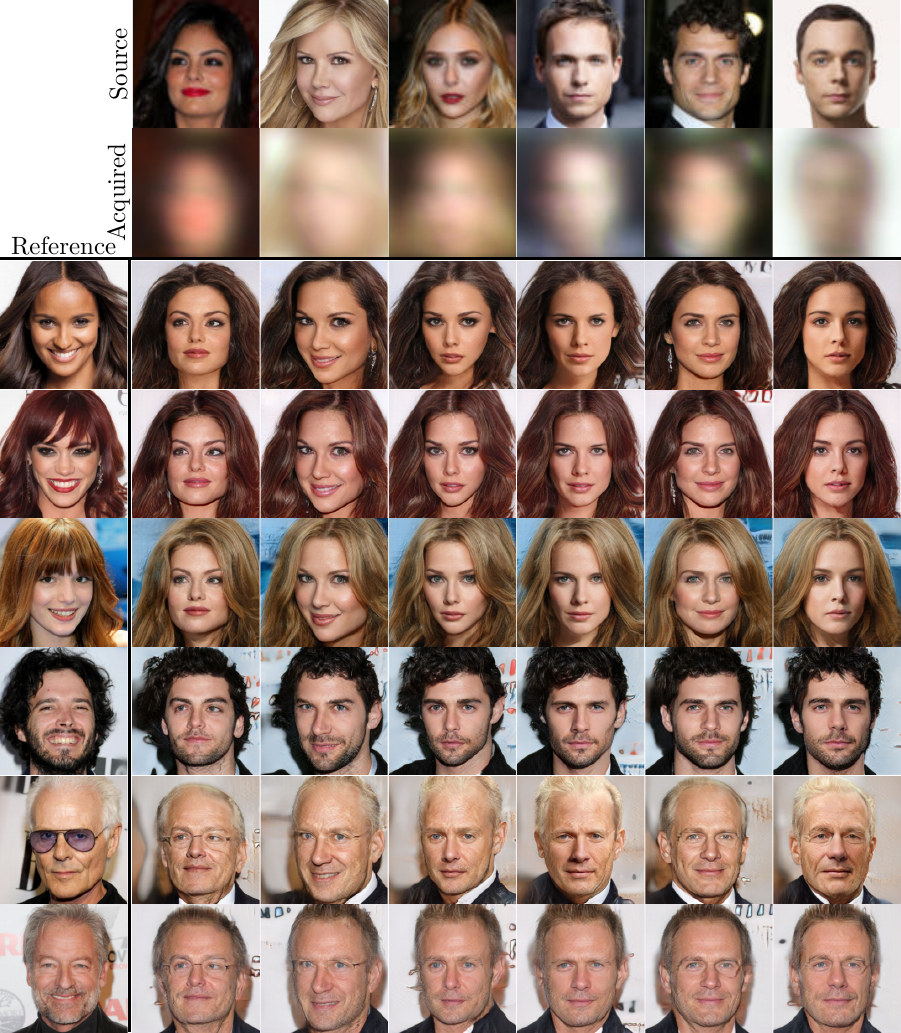}
    \caption{\small \textbf{Qualitative results}. We qualitatively evaluate our proposed face de-identification method using the CelebA-HQ dataset.}
    \label{fig:qualitative_rstl}
\end{figure}
\begin{figure}[!t]
    \centering
    \includegraphics[width=\columnwidth]{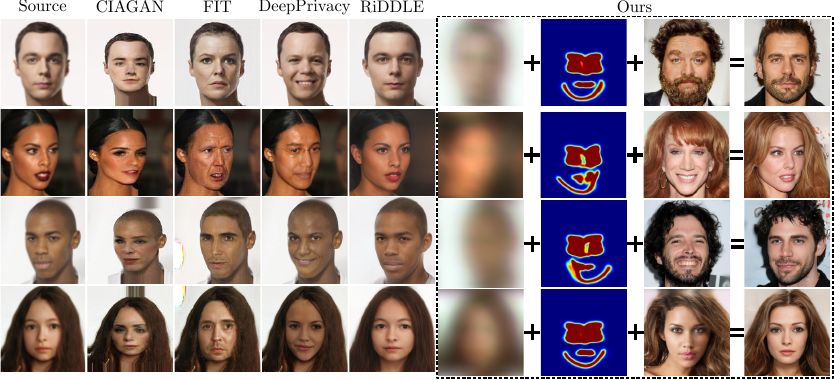}
    \vspace{-15pt}
    \caption{\textbf{Qualitative Results}. We compare our method against SOTA methods: CIAGAN, FIT, DeepPrivacy, and RIDDLE.}
    \label{fig:celeba_rstl}
    \vspace{-10pt}
\end{figure}

\mysection{Qualitative Results}. Figure \ref{fig:qualitative_rstl} shows generated faces using our proposed framework. In the first row, we show the source image, which is the input of our privacy-preserving camera. The image acquired by our camera is shown in the second row. Our generative network generates new faces using the acquired privacy-preserving image and different reference images shown in the first column. We observe that the generated images have the style extracted from the reference images and face geometry of the source images. Furthermore, we qualitatively compare our approach with SOTA methods in Fig.~\ref{fig:celeba_rstl} using the CelebA-HQ dataset. While our generated faces might not achieve the same level of realism as those produced by methods like RiDDLE, they effectively fulfill our primary goal of preserving privacy.
%


\subsection{Robustness to Deconvolution}
\label{sec:deconv}

\begin{figure}
    \centering
    \includegraphics[width=\columnwidth]{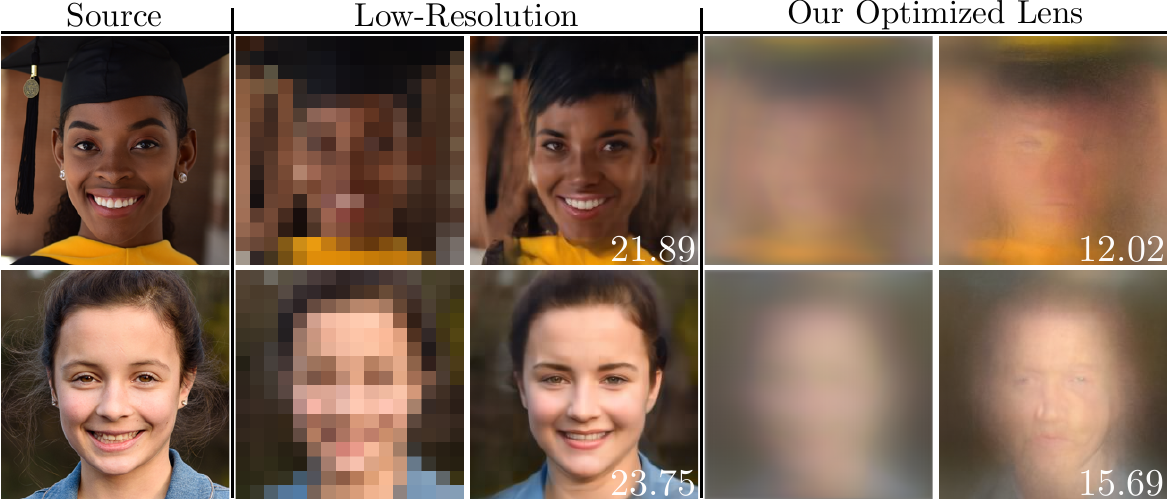}
    \caption{\small \textbf{Robustness to Deconvolution.} We use the SOTA diffusion model in \cite{kawar2022denoising} to recover the person identity. We show the PSNR between the original and recovered face in each image.}
    \label{fig:diff_rstl}
    \vspace{-14pt}
\end{figure}

Assuming that an attacker performs an adversarial attack and successfully captures the privacy-preserving image (see Fig. \ref{fig:intro} (b)). Then, the attacker could use a deconvolution method to recover the identity of the person. To test the robustness of our optimized lens to deconvolution attacks, we use a SOTA diffusion model (DDRM) proposed by Kawar et al.  \cite{kawar2022denoising} to recover the original image. DDRM is trained to recover images in multiple scenarios, such as superresolution, inpainting, colorization, and deblurring. We apply the pre-trained DDRM model on privacy-preserving images captured by our optimized lens and a low-resolution camera with $16 \times 16$ sensor size. As shown in Fig. \ref{fig:diff_rstl}, DDRM can recover the identity of the person from the low-resolution with high accuracy but fails to recover the face from the privacy-preserving image acquired by our optimized lens. Similar to previous works \cite{hinojosa2021learning, hinojosa2022privhar}, we also investigate the case where an attacker has direct access to the camera; hence, it can get the PSF by imaging a point of source light and perform the attack using non-blind deconvolution methods (\eg, the Wiener Filter). See our supplementary document for results on non-blind deconvolution attacks.

\subsection{Human Evaluation for Privacy Protection}

We conduct a human evaluation to assess the effectiveness of privacy protection given by our approach under four different scenarios. In the first scenario,  we randomly select five original face images alongside their corresponding ``low-resolution'' and ``blurred''  image versions. Here, the term ``blurred'' refers to the privacy-preserving images acquired through our optimized lens; then, we display the images to 106 individuals and ask them to judge to what degree they believe that the identity of the person is protected considering the following options: \textit{Not Protected}, \textit{Slightly Protected}, \textit{Moderately Protected}, and \textit{Highly Protected}. We found that only $12.76 \%$ of the surveyed people consider that the ``low-resolution'' face image is \textit{Highly Protected} while $56.00\%$ consider that the ``blurred'' face image is \textit{Highly Protected}. In the second scenario, we generate a new face image with our face de-identification framework and show it alongside the other five face images, where one of them corresponds to the current person we want to protect (source image). Then, we asked the 106 individuals to identify the source image. The responses show that $91.82\%$ of the people fail to identify the source image, showing the effectiveness of our method. Refer to our supplementary document for details about our human evaluation study.




\subsection{Hardware Experiments}

\begin{figure}
 \centering
  \includegraphics[width=\columnwidth]{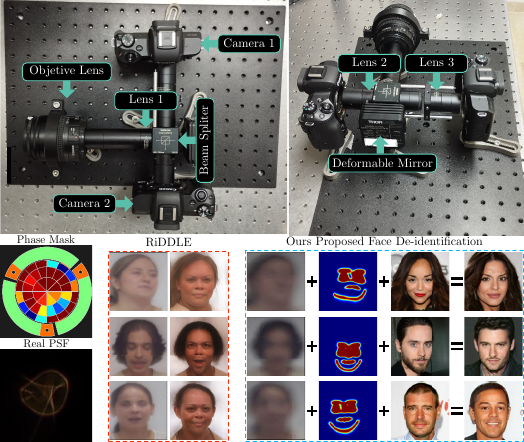}
 \caption{\textbf{Proof-of-concept optical system}. (Top) Experimental hardware setup for our privacy-preserving approach. (Bottom) Qualitative results on images acquired by our prototype camera and comparison with the RiDDLE method.}
 \vspace{-12pt}
 \label{fig:lab_mnt}
\end{figure}

We developed a real camera prototype in the optical laboratory to validate our proposed approach; see Fig. \ref{fig:lab_mnt} (top). Our prototype includes an additional side information branch to obtain the original images without privacy protection, which are used as ground truth to estimate DIS metrics. We emulate the lens designed with our framework using a deformable Thorlabs mirror (DMH40-P01)  in a 4f optical system built with 50mm achromatic lenses and two Canon EOS M50 mirrorless cameras as sensors. Due to the deformable mirror limitations, we are restricted to using only $p=15$ Zernike coefficients. Therefore, considering this limitation, we first simulate the first stage in our proposed framework (Sec. \ref{sec:stageI}) to obtain the simulated coefficients. Then, we load the learned coefficients to the deformable mirror and calibrate the optical setup. After calibration, we obtained the following learned Zernike coefficients: $\{\beta_1, \beta_2, \beta_3 = 0.0, \beta_4 = -0.83, \beta_5 = 0.0, \beta_6 = -0.31, \beta_7 = 0.08, \beta_8 = -0.69, \beta_9 = 0.0, \beta_{10} = 0.0, \beta_{11} = 0.0, \beta_{12} = -0.67, \beta_{13} = 0.0, \beta_{14} = 0.18, \beta_{15} = 0.0 \}$. We acquired 17 short face videos, each lasting between 15 to 30 seconds, to capture the same person with different facial expressions and head positions. After preprocessing, we obtain $3700$ face images, which we split into two sets of $3000$ and $700$ images for finetuning and testing, respectively. Specifically, we finetune the Heatmap Regression Network $\mathcal{U}$ and the Generator $\mathcal{G}$ of our model to adapt them to our acquired dataset. To evaluate the effectiveness of our proposed method on the data obtained in our lab, we use the DIS metric to realize a comparison with the SOTA RiDDLE model \cite{li2023riddle} over the collected testing set, see Tab. \ref{tab:lab_rstl}. Furthermore, we also show some qualitative comparisons in Fig. \ref{fig:lab_mnt} (bottom). These visual and quantitative results demonstrate that our low-resolution approach and proposed learned camera are effective for face anonymization in real-world scenarios. Please refer to our supplementary material for a qualitative comparison when using a low-resolution camera with data acquired in our lab.


\begin{table}[t]
\centering
\resizebox{0.65\columnwidth}{!}{%
\begin{tabular}{cccc}
\hline
\multirow{2}{*}{Method} & \multicolumn{3}{c}{DIS$\uparrow$}                \\ \cline{2-4} 
                        & FR             & CASIA          & VGGFace2       \\ \hline
RiDDLE                  & {\ul 0.798}    & 0.958          & 1.224          \\
Ours-LR                  & \textbf{0.808} & {\ul 1.141}    & {\ul 1.310}    \\
Ours                & 0.792          & \textbf{1.156} & \textbf{1.334} \\ \hline
\end{tabular}%
}
\vspace{-5pt}
\caption{\textbf{Quantitative results - Prototype camera.} We quantitatively evaluate the performance of our proposed approach using the privacy-preserving data acquired by our prototype camera and compare it against RiDDLE, which uses traditional images.}
\vspace{-12pt}
\label{tab:lab_rstl}
\end{table}




\section{Conclusion}
\label{sec:conclusions}

We presented a novel hardware-level face de-identification approach that proposes a solution to close the security gap between image acquisition and algorithms in traditional software-level methods. To ensure training stability and efficiency, we propose a two-stage optimization process. First, we leverage optics learning to capture privacy-preserving images while preserving facial features for face heatmap regression. Then, we employ a generative adversarial network to generate new synthetic faces that preserve facial expression from the source image but style and face appearance from a reference image with no privacy concern. Specifically, our proposed approach offers two key advantages over existing methods as \textbf{Hardware-Level Security} mitigating the vulnerability to sniffing attacks that can compromise privacy protection, and \textbf{High-Quality Face Generation} using our proposed generative framework for privacy-preserving images acquired with our camera. 

\mysection{Limitations.} In our prototype camera, the deformable mirror is the main limitation, which only uses 15 Zernike Polynomials, limiting the scene's level of distortion. Additionally, generating faces using only our privacy-preserving images is still challenging for the generative model; hence, some artifacts could appear in the generated faces.

\mysection{Acknowledgment.} This work was supported by the King Abdullah University of Science and Technology (KAUST) Office of Sponsored Research through the Visual Computing Center (VCC) funding, the SDAIA-KAUST Center of Excellence in Data Science and Artificial Intelligence (SDAIA-KAUST AI), and the Universidad Industrial de Santander (UIS), Colombia under the research projects VIE-3735 and VIE-3924.



{
    \small
    \bibliographystyle{ieeenat_fullname}
    \bibliography{main}
}


\end{document}